\documentclass[runningheads]{llncs}
\usepackage{graphicx}
\usepackage{booktabs, multirow} % for borders and merged ranges
\usepackage{soul}% for underlines
\usepackage{lineno,hyperref,graphicx}
\usepackage{svg}
\usepackage{amsmath}
\usepackage{amssymb}
\usepackage{caption}
\usepackage{subcaption}
\usepackage{caption} 
\begin{document}
\title{Probing Semantic Grounding in Language Models of Code with Representational Similarity Analysis}
% \author{Shounak Naik \\\inst{1}\orcidID{0000-1111-2222-3333} \and
% Rajaswa Patil\inst{2,3}\orcidID{1111-2222-3333-4444} \and
% Swati Agarwal\inst{3}\orcidID{2222--3333-4444-5555}}

\author{Shounak Naik\inst{1,2,3} \and
Rajaswa Patil\inst{1,2,3} \and
Swati Agarwal\inst{1,2}\orcidID{0000-0001-9586-2794} \and
Veeky Baths\inst{1,3}\orcidID{0000-0001-9980-5738}}
\authorrunning{S. Naik et al.}
\institute{BITS Pilani, K. K. Birla Goa Campus, India \and
Computational Linguistics and Social Networks Lab, BITS Pilani, Goa, India \and
Cognitive Neuroscience Lab, BITS Pilani, Goa, India\\
\email{\{f20170835, f20170334, swatia, veeky\}@goa.bits-pilani.ac.in}
}
\maketitle  % typeset the header of the contribution
\begin{abstract}
Representational Similarity Analysis is a method from cognitive neuroscience, which helps in comparing representations from two different sources of data. In this paper, we propose using Representational Similarity Analysis to probe the semantic grounding in language models of code. We probe representations from the CodeBERT model for semantic grounding by using the data from the IBM CodeNet dataset. Through our experiments, we show that current pre-training methods do not induce semantic grounding in language models of code, and instead focus on optimizing form-based patterns. We also show that even a little amount of fine-tuning on semantically relevant tasks increases the semantic grounding in CodeBERT significantly. Our ablations with the input modality to the CodeBERT model show that using bimodal inputs (code and natural language) over unimodal inputs (only code) gives better semantic grounding and sample efficiency during semantic fine-tuning. Finally, our experiments with semantic perturbations in code reveal that CodeBERT is able to robustly distinguish between semantically correct and incorrect code.

\end{abstract}
\section{Introduction}

Recent development in deep neural network-based (DNN) language modeling has brought in great advancements in various data-driven artificial intelligence (AI) technologies. With the increasing scale of data repositories, the usage of language models in various data-driven AI applications has increased as well. This has brought in a new paradigm of \emph{Language-Model-as-a-Service} \cite{blackbox2022sun}. Under this paradigm, language models trained on huge web data are available as cloud platforms, and enable natural language interfaces to various AI applications, as well as provide rich features for various downstream applications such as sentiment analysis, question answerin, text completion, etc.

\begin{figure}[!htbp]
  \centering
  \includegraphics[width=\columnwidth]{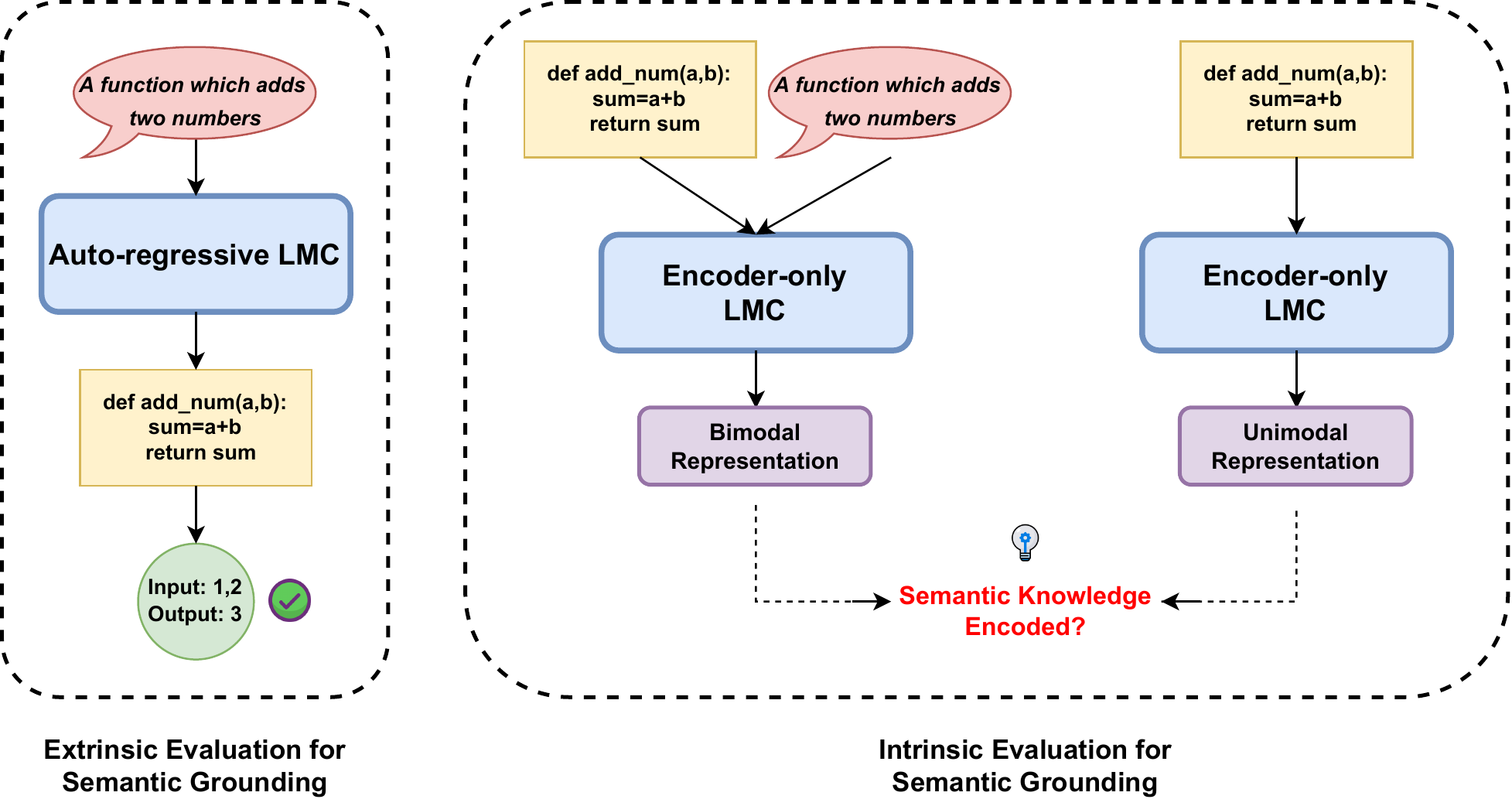}
  \caption{Difference between extrinsic and intrinsic evaluation for semantic grounding in language models of code (LMCs). Encoder-only LMCs can produce Bimodal representations (left) or Unimodal representations (right) depending on the type of input.}
  \label{fig:evaluation-type}
\end{figure}

Recent efforts in building AI-based assistive technologies in code and software has seen development of big code datasets and repositories  \cite{pile,codesearchnet,puri2021codenet}. Hence, in parallel to the progress in developing DNN language models for natural language, there has been a recent spurt in developing language models for code (LMC) \cite{feng-etal-2020-codebert,ahmad-etal-2021-unified,chen2021codex,wang-etal-2021-codet5}. Representations obtained from LMCs are used as features for a variety of downstream applications such as code search, code translation, code refinement, bug detection, data wrangling, querying databases, etc. Traditionally, the representations from language models are either evaluated against downstream tasks (extrinsic evaluation), or are probed for specific knowledge (intrinsic evaluation). A robust language model of code is ideally expected to capture the actual meaning of the code text (semantics) and not just its surface level form-based statistical patterns. As the underlying code repository and databases become large and complex, semantic grounding becomes quite important to avoid spurious outcomes.

Auto-regressive LMCs like Codex (GPT-3) \cite{chen2021codex} can be probed for such semantic grounding with code-generation tasks as shown in Figure~\ref{fig:evaluation-type}. This can either be achieved by generating code for a given natural language description and testing it against input-output (IO) specifications \cite{austin2021program}, or by generating the correct return sequences for a given function \cite{chen2021codex}. These methods though complete from a semantic evaluation perspective, pose several limitations. Firstly, these are extrinsic evaluation methods which only focus on generating form-level tokens, which does not guarantee true semantic grounding \cite{bender-koller-2020-climbing}. Secondly, for tasks not involving code-generation and for the tasks that involve scaling to large code databases, the representations obtained from a LMC are more important than the its ability to generate code and text (Figure~\ref{fig:evaluation-type}). Hence, intrinsic evaluation methods become necessary for evaluating the semantic grounding under such settings. Finally, for encoder-only models like CodeBERT \cite{feng-etal-2020-codebert}, code-generation is not technically feasible, which is a requirement for testing semantic grounding against IO specifications.

In this work we propose using Representational Similarity Analysis (RSA) \cite{rsa2008} to probe the semantic grounding in the representations from LMCs. Unlike the other previous intrinsic evaluation methods, RSA is a method from cognitive neuroscience which offers a flexible intrinsic evaluation setting which is agnostic to various properties of the representations like their source, size, structure, modality, etc. Hence, RSA becomes a good choice to evaluate the representations obtained from LMCs against a ground truth semantic representation. In this work we probe CodeBERT with RSA. We use a code's natural language description as its ground truth semantic representation. Through our experiments we show how RSA can be utilized to evaluate various aspects of research and development in using LMCs for various data-driven code intelligence tasks. Specifically, we aim to use RSA-based semantic grounding evaluation to study the following:

\begin{enumerate}
 \item The localization of semantic knowledge in different layers of CodeBERT.
 \item The impact of size of the fine-tuning datasets on the semantic grounding in CodeBERT.
 \item The impact of modality of code repositories (unimodal vs. bimodal) on the semantic grounding in CodeBERT.
 \item The robustness of CodeBERT model against semantic perturbations in the code.
\end{enumerate}

\section{Background}
\label{sec:background}

% \paragraph{Language Models of Code:}
% LMC general
Recent progress in deep learning based transformer language modeling has resulted in development of various language models of code (LMCs). LMCs can be broadly classified into two categories: auto-regressive models and encoder-only models. Auto-regressive models like PL-BART \cite{ahmad-etal-2021-unified}, Codex (GPT-3) \cite{chen2021codex}, and CodeT5 \cite{wang-etal-2021-codet5} are pre-trained to generate code and text. Whereas, encoder-only models like CodeBERT \cite{feng-etal-2020-codebert} are pre-trained to encode representations of code, and do not generate any code or text. Depending on their category and their neural architecture, the LMCs are pre-trained with a variety of pre-training objectives by using huge multimodal (natural language and programming language) datasets \cite{pile,codesearchnet,puri2021codenet}.

% LMC applications (to build relevance to venue)
LMCs have various data-driven applications in programming and software engineering. These include natural language based code search \cite{codesearchnet}, code translation and refinement \cite{codexglue}, code repair and bug detection \cite{codexglue}, and natural language interfaces to databases \cite{yu-etal-2018-spider}. Representations from pre-trained LMCs are used to enable such applications in practical deployments. LMCs are usually fine-tuned with a relatively smaller dataset for the target downstream application before deployment. This process is usually quite resource-heavy in terms of data annotation costs and fine-tuning computing costs. Hence, getting an estimate of optimal amount of fine-tuning for robust deployments is very important. LMCs are usually evaluated against standard tasks and benchmarks \cite{codexglue,codesearchnet}. While most such recent LMCs show great performance on such benchmarks, the true code understanding capabilities of these models are still being tested with semantically challenging extrinsic evaluation settings \cite{austin2021program,chen2021codex}. We extend such semantic evaluations into the intrinsic evaluation paradigm by probing the representations from these LMCs for semantic grounding.

% \paragraph{Representational Similarity Analysis:}
% RSA general
Representational Similarity Analysis (RSA) is a method from cognitive neuroscience, originally invented to compare representations of neural and physiological data and signals from different sources \cite{rsa2000,rsa2008}. Recent work in natural language processing research has focused on using RSA for various interpretability studies with language models. Previously, RSA has been used to ground neural representations from language models to that in the human brain \cite{gauthier-levy-2019-linking}. RSA has also been used to probe contextual semantic and syntactic information in language models \cite{abnar-etal-2019-blackbox,lepori-mccoy-2020-picking}. RSA has also been proved to be quite useful in studying the effect of fine-tuning on natural language models \cite{merchant-etal-2020-happens}. Given the versatility of the RSA method, and its successful application in evaluating natural language models, it becomes a great choice to perform intrinsic evaluation of LMCs.

\section{Experimental Setup}
\label{sec:setup}

In this section we describe the setup required for our experiments. We discuss the details of the dataset used in our experiments in Section`\ref{sec:dataset}. Whereas, the description of our RSA-based probing approach is explained in Section~\ref{sec:rsa}. Finally, we discuss our experiments and their outcomes in Section~\ref{sec:experiments}.

\subsection{Dataset}
\label{sec:dataset}

We use the IBM CodeNet dataset \cite{puri2021codenet} for all our experiments. The dataset comprises $4000$ coding problems with submissions in multiple programming languages. The problem descriptions in the dataset can be used as a natural language (NL) modality. Whereas, the code submissions to these problems can be used as a programming language (PL) modality. Hence each sample in the dataset can be viewed as a NL-PL pair. The dataset is available in the form of web files (HTML) for problem descriptions and spreadsheet files for problems and code submissions related metadata. While the original CodeNet dataset comprises of code samples from over $50$ programming languages, we only focus on the six programming languages supported by CodeBERT as shown in Table~\ref{tab:data-stats}. We sub-sample and clean the CodeNet dataset as per the requirements for our RSA-based semantic grounding probing experiments.

\begin{table}[thbp]
\centering
\resizebox{0.9\columnwidth}{!}{
\begin{tabular}{@{}cccccccc@{}}
\toprule
\textbf{\begin{tabular}[c]{@{}c@{}}Data\\ Split\end{tabular}} & \textbf{\begin{tabular}[c]{@{}c@{}}Submission\\ Type\end{tabular}} & \textbf{{Go}} & \textbf{{Java}} & \textbf{{JS*}} & \textbf{{PHP}} & \textbf{{Python}} & \textbf{{Ruby}} \\ \midrule
\multirow{2}{*}{\textbf{Test}}  & \textbf{Incorrect} & 10482 & 22213& 8846 & 6339& 25322 & 15354\\
& \textbf{Correct}& 22768 & 25438& 9729 & 8103& 25500 & 23658\\ \midrule
\textbf{Training}& \textbf{Correct}& 29557 & 62370& 12377& 10455  & 67337 & 45433\\ \bottomrule
\end{tabular}
}
\caption{The number of NL-PL sample pairs across the six programming languages used by CodeBERT in the final dataset. \footnotesize{*JS: JavaScript}}
\label{tab:data-stats}
\end{table}

To generate the test data, we filter out $255$ problems such that each problem has a correct submission, and a wrong submission in each of the six programming languages under consideration: Go, Java, JavaScript, PHP, Python, and Ruby. In order to generate the training data, we filter out $808$ problems ($708$: Training, $100$: Validation), such that each problem has at least one correct submission for each of the six programming languages. Given the noisy metadata files from the CodeNet dataset, we manually extract the problem descriptions from the raw problem description web files in the dataset. The extracted set of problem descriptions is multilingual (non-English descriptions) in nature. We translate all the problem descriptions to English using an off the shelf translation tool: DeepL.\footnote{\href{https://www.deepl.com/translator}{https://www.deepl.com/translator}} The translations obtained from DeepL are manually checked for any errors. The final dataset statistics are shown in Table~\ref{tab:data-stats}. The test data split is used to perform the RSA-based probing for semantic grounding in CodeBERT. On the other hand the training data split is used to fine-tune the CodeBERT model, in order to inspect the role of fine-tuning on its semantic grounding. The original IBM CodeNet dataset can be found \href{https://developer.ibm.com/exchanges/data/all/project-codenet/}{here}. Our cleaned and modified version of the IBM CodeNet dataset will be available on request.

\begin{figure}
    \centering
    \includegraphics[width=0.9\columnwidth]{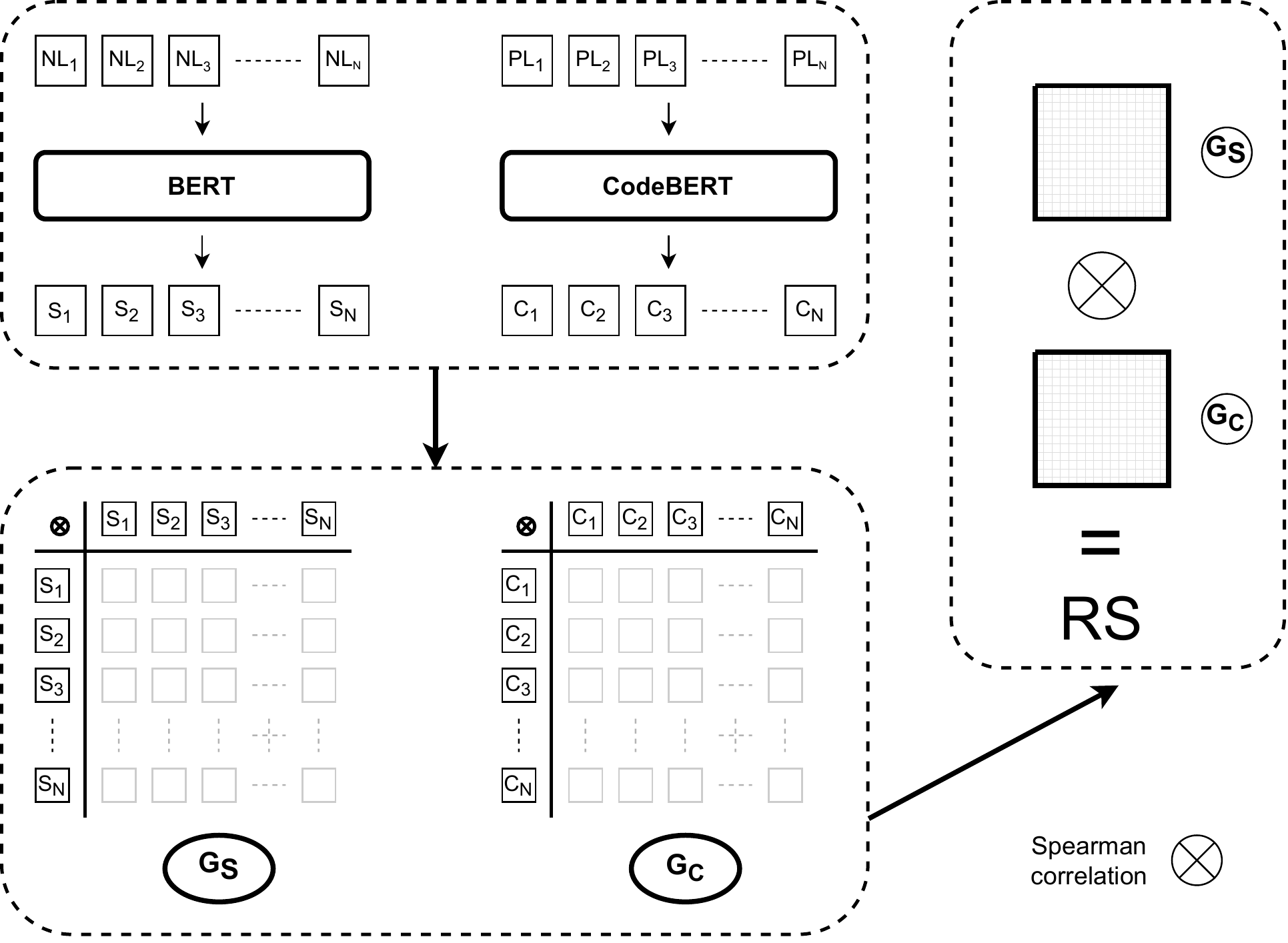}
    \caption{Schematic overview of using the Representational Similarity Analysis method with the code representations from the CodeBERT model.}
    \label{fig:rsa}
\end{figure}

\subsection{Representational Similarity Analysis (RSA)}
\label{sec:rsa}

Given $N$ natural language description - code snippet pairs (NL-PL): we first obtain the code representations $\{C_k\}$ from the code snippets with CodeBERT. Similarly, we obtain the semantic representations $\{S_k\}$ from the natural language descriptions with BERT, where $k \epsilon [1,N]$. For ablation purposes, we extract code representations under two different input settings: Unimodal (PL-only) and Bimodal (NL and PL) as supported by the CodeBERT model. Next, we construct the individual representational geometries ($\mathcal{G} \in \mathbb{R}^{N \times N}$) for $\{C_k\}$ and $\{S_k\}$ by computing the pairwise dissimilarities between all the samples in the dataset:

\begin{equation}
 \mathcal{G}_{C} = \{1-similarity(C_i, C_j)\} 
\end{equation}

\begin{equation}
 \mathcal{G}_{S} = \{1-similarity(S_i, S_j)\} 
\end{equation}

where, $i,j \epsilon [1,N]$

The final representational similarity score ($RS$) between the code and semantic representations can then be obtained by finding the similarity between the two representational geometries:

\begin{equation}
 RS_{(\mathcal{C}, \mathcal{S})} = similarity(\mathcal{G}_{C}, \mathcal{G}_{S})
\end{equation}

A higher $RS$ value can then be interpreted as higher semantic grounding in the representations of code. We use Spearman correlation coefficient as the $similarity$ measure while calculating the pair-wise dissimilarities between the samples, as well as the similarities between the two representational geometries. We obtain statistically significant similarity scores throughout our experiments with correlation p-values $< 0.05$. A schematic overview of using RSA with representations from the CodeBERT model is shown in Figure~\ref{fig:rsa}. While Figure~\ref{fig:rsa} shows Unimodal setting with CodeBERT (PL-only), we perform RSA with Bimodal setting as well (NL and PL).

\begin{figure}[!htbp]
  \centering
  \includegraphics[width=\columnwidth]{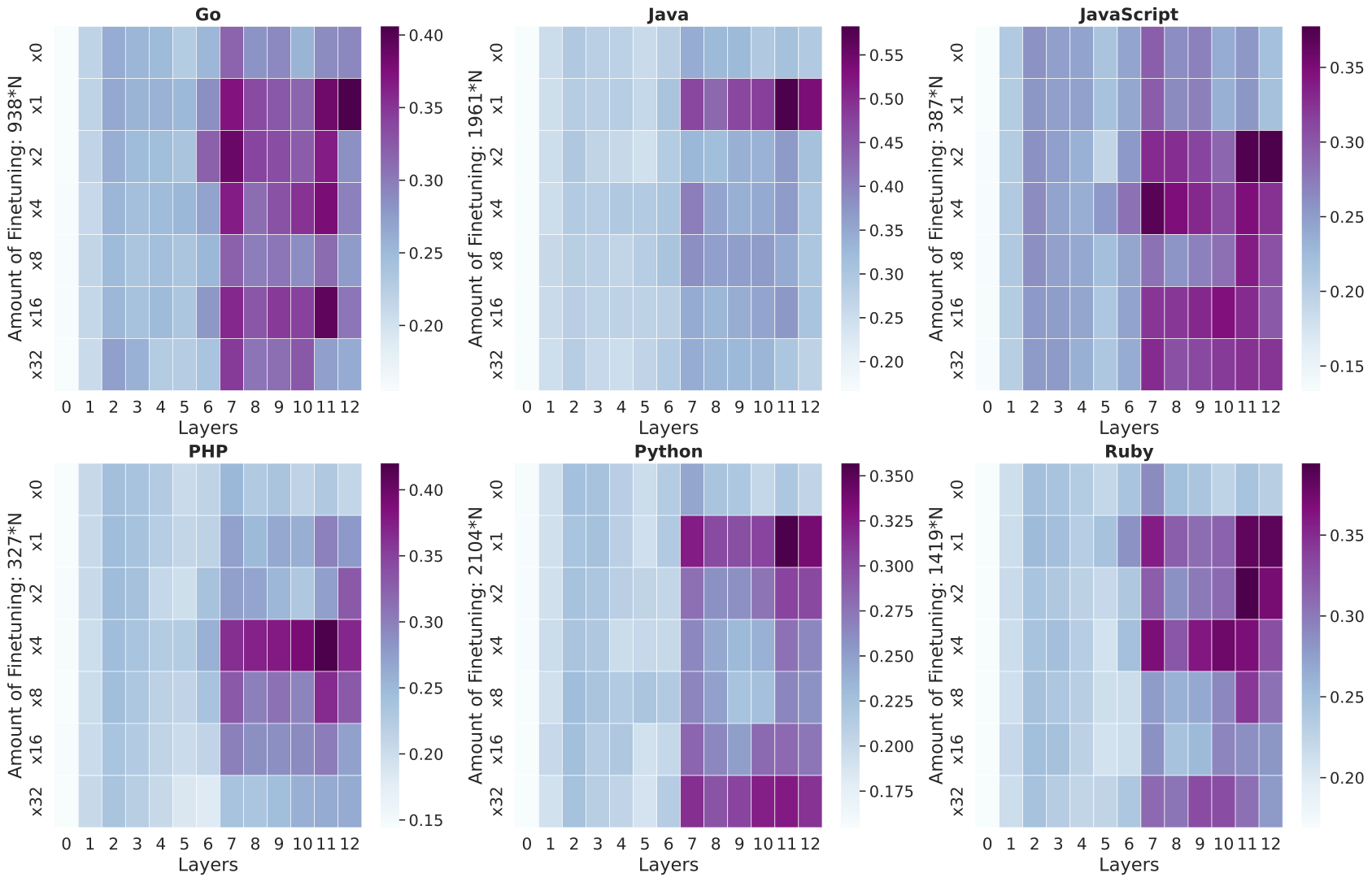}
  \caption{Heatmaps of RSA similarity scores across the six programming languages and $13$ layers of the CodeBERT model.}
  \label{fig:heatmap}
\end{figure}

\section{Experiments}
\label{sec:experiments}

Previous work in language model probing has tried to analyze various semantic knowledge localization trends in the layers of a language model \cite{rogers-etal-2020-primer}.\footnote{Localization studies aim to find out which layer and parameters of a neural network capture the maximum amount of specific knowledge.} This helps understand the model dynamics, as well as in selecting the best representations for downstream applications. Hence, we first begin by evaluating the semantic grounding in the pre-trained CodeBERT model with the default bimodal input setting. We perform RSA with all 13 representations\footnote{Embedding (layer 0) + Encoder blocks (layer 1-12)} from the pre-trained CodeBERT model. We use the \emph{Correct} code samples from the Test split of the data here.  We observe that the semantic grounding is consistently low for the pre-trained model (Figure~\ref{fig:heatmap} x0) across all the layers and programming languages with no particular localization trends. This indicates that the pre-trained CodeBERT model shows low levels of semantic understanding for code data, and directly using the representations from the pre-trained model in downstream practical applications might not yield robust performance. Taking this into consideration, we probe various practically motivated aspects of semantic grounding in LMCs:

\subsection{Semantic Fine-tuning}
 We start our experiments by evaluating layer-wise representational similarity scores for CodeBERT model. We visualize the extent of similarity and representative semantic grounding through heatmap plots as shown in Figure~\ref{fig:heatmap}. We plot a separate heatmap for each of the six programming languages used by CodeBERT. The X-axis of each heatmap represents the layer of the CodeBERT model. The Y-axis of each heatmap represents the amount of fine-tuning data used to induce semantic knowledge in the CodeBERT model. The heatmap gradient represents the RSA similarity score, which is representative of the amount of semantic grounding as discussed in Section~\ref{sec:rsa}.

 Firstly, we observe that the pre-trained CodeBERT model does not show significantly low semantic grounding across all the layers and programming languages (as seen in the first row of each heatmap). Since the original pre-training tasks used by the CodeBERT model do not induce enough semantic knowledge in its representations as seen in Figure~\ref{fig:heatmap}, we evaluate how the amount of downstream semantic fine-tuning affects the model's semantic grounding. We divide the training data of each programming language into six splits, where number of samples are increased in the power of $2$ at every step: x0, x1, x2, x4, x8, x16, and x32. We use the NL-PL pairs in the training data to fine-tune the model on the semantically relevant task of semantic code search - one of the major downstream applications of code representations. We observe that fine-tuning the model helps with inducing semantic grounding in the representations (Figure~\ref{fig:heatmap}). Even fine-tuning on a very small number of samples, significantly increases the semantic grounding. Most of the semantic grounding is localized in the deeper layers of the model (right half of the heatmaps), which is similar to that of previous natural language models \cite{rogers-etal-2020-primer}. We also observe that the semantic grounding peaks at the pre-final layer. This shows that using representations from the final-layer or from the pre-trained model might not be optimal for data-driven downstream code applications. This also reveals that large multimodal datasets of NL-PL code samples are not required to induce high levels of semantic grounding in the CodeBERT model.

 \begin{figure}[htbp]
 \centering
\includegraphics[width=\columnwidth]{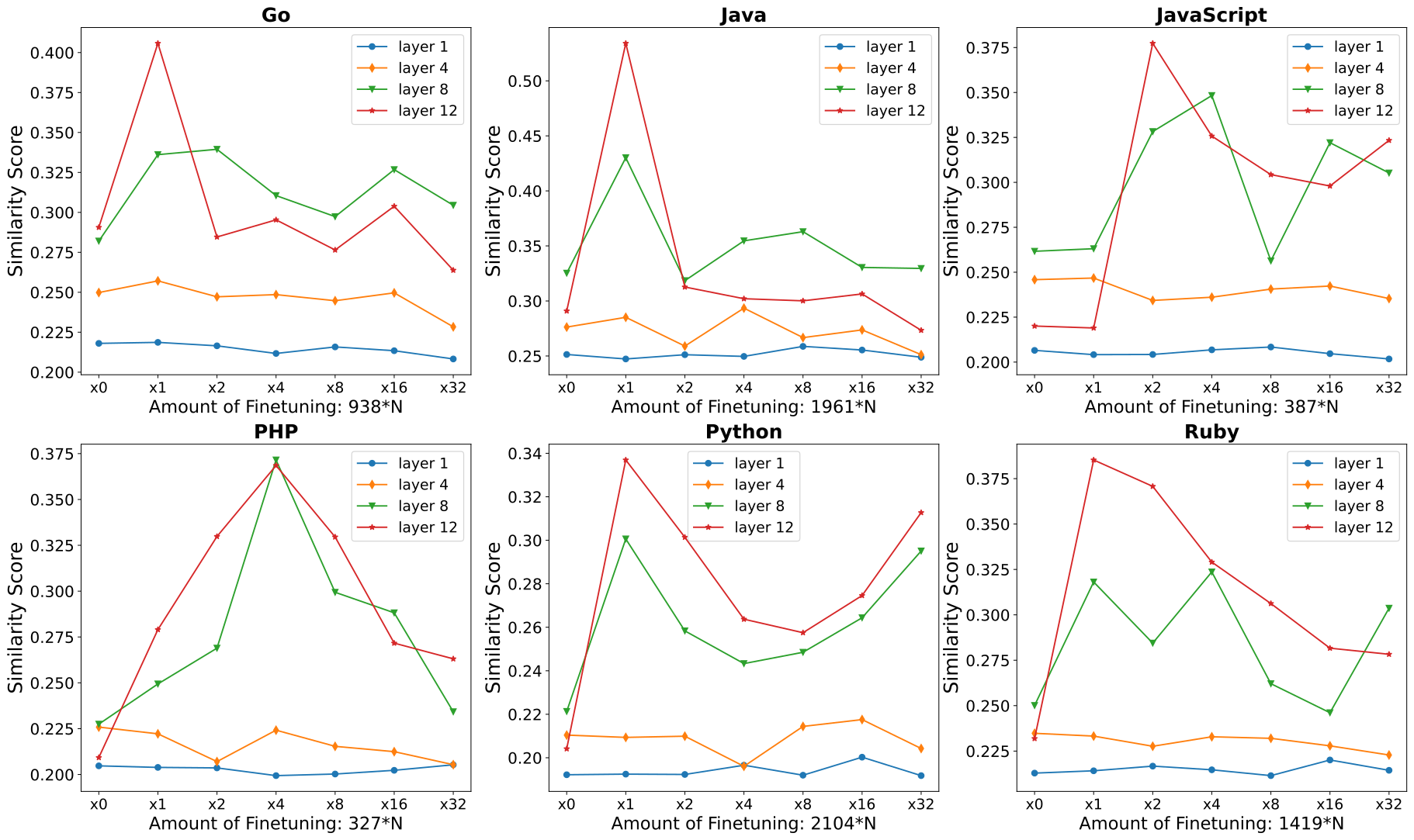}
\caption{RSA similarity scores after fine-tuning the CodeBERT model with Bimodal (NL and PL) input.}
\label{fig:nl-pl}
\end{figure}

\subsection{Input Modality}
\begin{figure}[htbp]
 \centering
\includegraphics[width=\columnwidth]{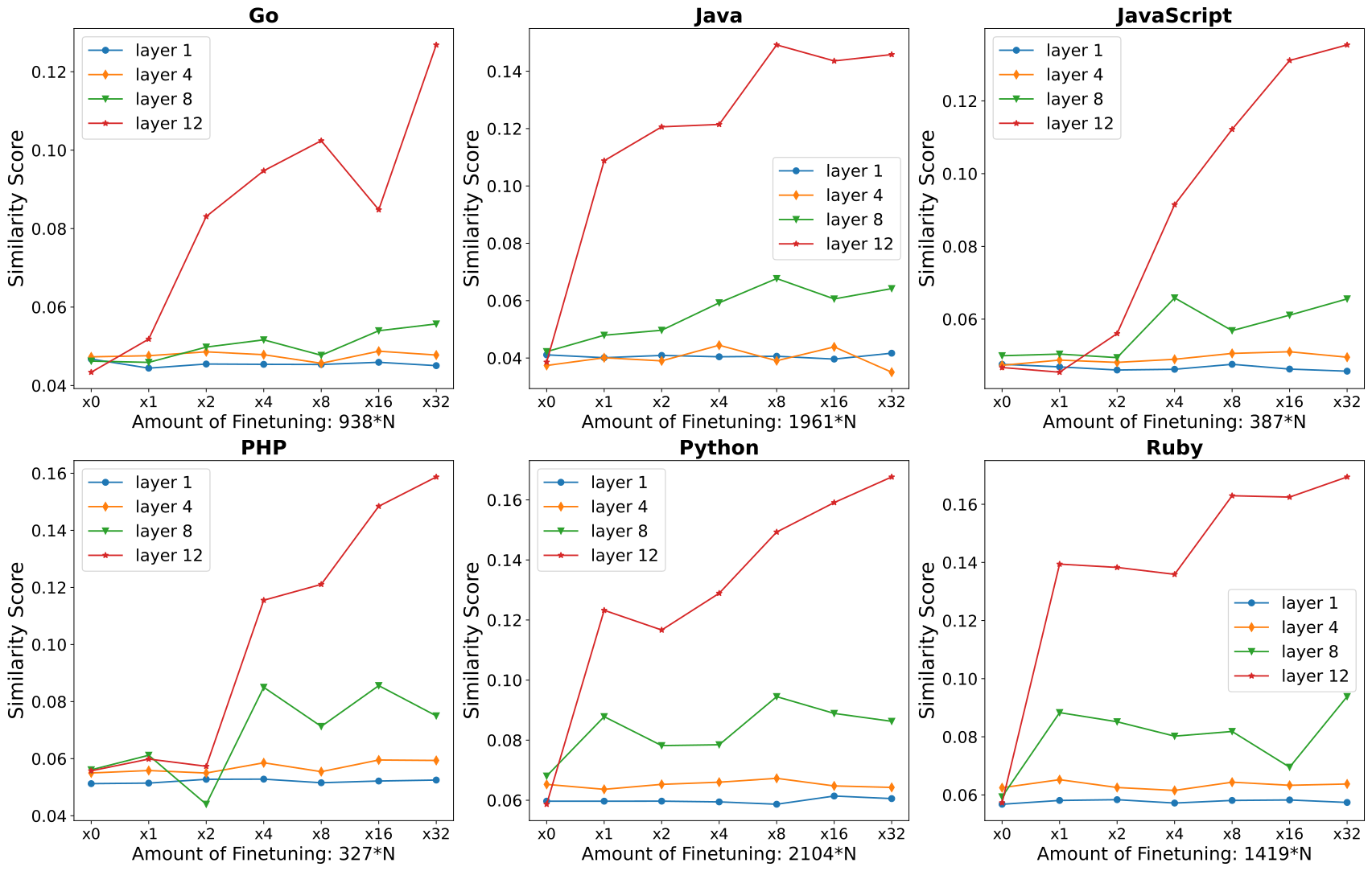}
\caption{RSA similarity scores after fine-tuning the CodeBERT model with Unimodal (PL-only) input.}
\label{fig:pl-only}
\end{figure}

Under practical settings, code repositories can either be unimodal or bimodal in nature. Hence, most LMCs support bimodal NL and PL inputs. Here, we inspect the role of input modality in semantic grounding. Following the fine-tuning done with bimodal data as described in Section~\ref{sec:experiments}, we repeat the fine-tuning process with unimodal data. We visualize the effect of input modality on the semantic grounding in CodeBERT model with line-chart plots shown in Figure~\ref{fig:nl-pl} and Figure~\ref{fig:pl-only}. We plot separate sub-plots for each of the six programming languages used by CodeBERT. The X-axis of each sub-plot represents the amount of fine-tuning data used to induce semantic knowledge in the CodeBERT model. The Y-axis of each sub-plot represents the similarity score, which is representative of the amount of semantic grounding as discussed in Section~\ref{sec:rsa}. We report the scores for four layers for each of the six programming languages: \{1, 4, 8, 12\} - each of which has a separate color-coded trajectory in each of the sub-plots.

We observe that both unimodal and bimodal inputs show increasing semantic grounding in representations with fine-tuning in the deeper layers (8 and 12) as seen in Figure~\ref{fig:pl-only} and Figure~\ref{fig:nl-pl} respectively. On the other hand, similar to earlier findings early layers (1 and 4) do not show any significant semantic grounding. Hence, even without natural language descriptions, the deeper layers of the model seem to capture code semantics up to some extent by just looking at the code text.

\begin{figure}[!tbp]
  \centering
  \begin{minipage}[b]{0.51\textwidth}
    \includegraphics[width=\textwidth]{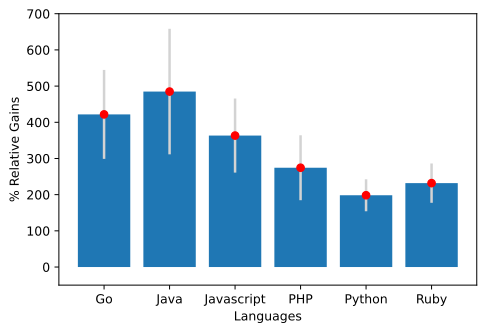}
    \caption{Amount of relative gains in semantic grounding when using bimodal inputs over unimodal inputs.}
    \label{fig:bimodal-vs-unimodal}
  \end{minipage}
  \hfill
  \begin{minipage}[b]{0.45\textwidth}
    \includegraphics[width=\textwidth]{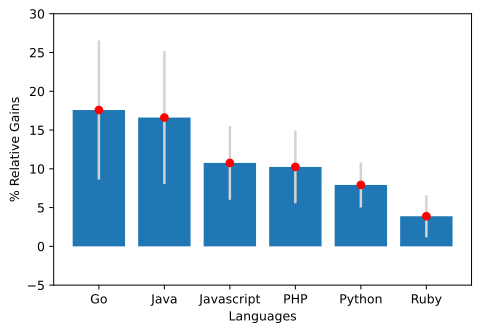}
    \caption{Amount of relative gains in semantic grounding when using \emph{Correct} code samples over \emph{Incorrect} code samples.}
    \label{fig:correct-incorrect-percent}
  \end{minipage}
\end{figure}

We also observe that representations from bimodal inputs hold significantly more semantic grounding (as high as 500\% in languages like Java) than those from unimodal inputs (Figure~\ref{fig:bimodal-vs-unimodal}). Hence, augmenting code repositories with natural language descriptions and comments can help in developing better downstream applications with LMC representations. Bimodal inputs also show better sample efficiency (Figure~\ref{fig:nl-pl}), where the performance peaks with significantly less amount of fine-tuning as compared to unimodal input which keeps on increasing with an increasing amount of fine-tuning data, while still showing lesser semantic grounding than bimodal input (Figure~\ref{fig:pl-only}).

\subsection{Semantic Perturbations}

While all our previous experiments are conducted with the \emph{Correct} code samples from the Test split of data, in this section we focus on using \emph{Incorrect} code samples which are semantically perturbed. Under practical settings, evaluating a code against test specifications provides a complete and strict evaluation of code semantics, where even a small change in code semantics shows error in the outputs. While this is possible with code generation models, intrinsically evaluating code representations is not possible with such a setting. In an attempt to enable such strict evaluation under an intrinsic setting, we compare the representational similarity scores for \emph{Correct} and \emph{Incorrect} submissions in the dataset under a unimodal (PL-only) setting. Using a unimodal setting ensures that cues from the natural language modality do not help the model capture semantics, and its true understanding of code semantics is tested. Overall, we observe that CodeBERT rightly shows significantly higher semantic grounding for \emph{Correct} submissions as compared to \emph{Incorrect} ones across all fine-tuning checkpoints and languages (Figure~\ref{fig:correct-incorrect-percent}). 

\begin{figure}[!htbp]
  \centering
  \includegraphics[width=\columnwidth]{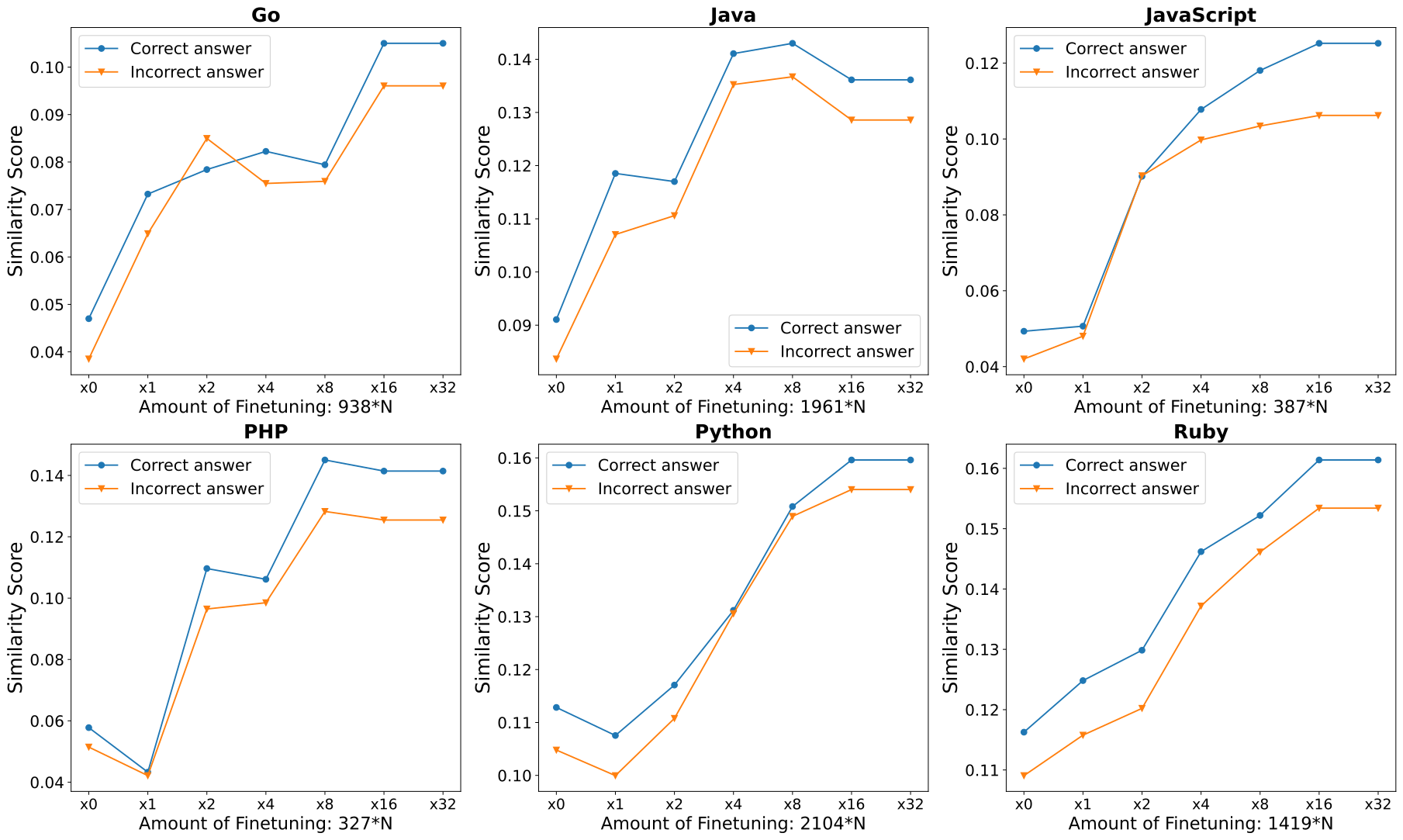}
  \caption{RSA similarity scores for \emph{Correct} and \emph{Incorrect} code samples obtained using Unimodal (PL-only) representations from the best-performing layer (11) of the CodeBERT model.}
  \label{fig:best-layer}
\end{figure}

Further, we inspect the effect of fine-tuning on the similarity scores for both \emph{Correct} and \emph{Incorrect} submissions. Here, we use unimodal representations from the best performing layer $11$ (Figure~\ref{fig:heatmap}). We visualize the semantic grounding trends in CodeBERT model for \emph{Correct} as well as \emph{Incorrect} submissions with line-chart plots as shown in Figure~\ref{fig:best-layer}. We plot separate sub-plots for each of the six programming languages used by CodeBERT. The X-axis of each sub-plot represents the amount of fine-tuning data used to induce semantic knowledge in the CodeBERT model. The Y-axis of each sub-plot represents the similarity score. 

\begin{figure}[htbp]
  \centering
  \includegraphics[width=0.6\columnwidth]{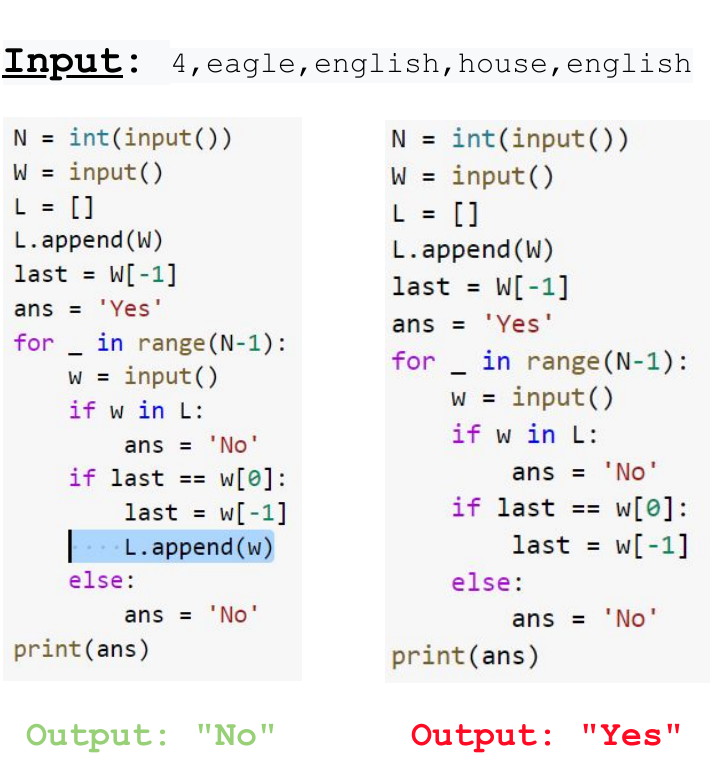}
  \caption{Example for \emph{Correct} (left) and \emph{Incorrect} (right) code samples from the CodeNet dataset with minor surface-level difference (highlighted) in form. \footnotesize{(Problem ID: p03261 ; Submission ID: s272963690 (left) s331881266 (right)}}
  \label{fig:code-example}
\end{figure}

We observe that with fine-tuning, the semantic grounding unexpectedly increases equally for both \emph{Correct} and \emph{Incorrect} submissions. This might suggest that similar to other language models, CodeBERT might be optimizing the code search task stochastically on surface-level forms, and not on the code meaning \cite{bender-koller-2020-climbing}. For example, Figure~\ref{fig:code-example} shows a \emph{Correct} and an \emph{Incorrect} submission to the same problem. Both the submissions only differ on a single line of code perturbation (highlighted), but produce semantically different outputs.  While active research is still trying to overcome such issues, this can be bypassed by using stricter ground-truth semantic representations derived from code structures like control flow graphs, data flow graphs, etc. and function specifications. Overall, CodeBERT consistently achieves more semantic grounding for \emph{Correct} submissions over the \email{Incorrect} submissions across all the six programming languages and amounts of fine-tuning data. This suggests that while CodeBERT is unable to penalize \emph{Incorrect} code samples for semantic perturbations, it is able to robustly differentiate between semantically correct and perturbed code samples by assign relatively lower semantic grounding to the \emph{Incorrect} samples. This forms an intrinsic evaluation alternative to the standard practice of evaluating against test specifications in an extrinsic evaluation setting.

\section{Conclusion and Future Directions}
\label{sec:conclusion}
In this work we propose using Representational Similarity Analysis to probe the semantic grounding in language models of code (LMC). Through our experiments with the pre-trained CodeBERT model we show that current pre-training methods do not induce semantic grounding in LMCs. We also show that fine-tuning on semantically relevant tasks helps induce semantic grounding in LMCs, which is localized in the deeper layers of a LMC. Overall, our experiments show that the representations from the pre-final layer of a LMC are most rich in semantic knowledge.  Our ablations with the input modalities reveal that different modalities of inputs show different types of semantic grounding and sample efficiency in LMCs, where even a relatively small number of fine-tuning examples is enough to obtain semantically robust performance in downstream applications. Our experiments with semantically perturbed \emph{Incorrect} code samples show that CodeBERT optimizes on form, and not on the meaning. However, it robustly assigns relatively lower semantic grounding to the \emph{Incorrect} code samples as compared to the \emph{Correct} ones across various programming languages and experimental settings.

While in the current work we use natural language descriptions as the ground truth semantic representation, in future more formal semantic representations like program structure and specifications can be used as the ground truth semantic representations. It is also quite imperative to study how insights obtained from RSA-based probing for semantic grounding translates to performance in practical deployments of downstream applications.

\section{Reproducibility}
We perform all our experiments on NVIDIA RTX A2000 GPU and Intel(R) Xeon(R) Gold 5120 CPU. In order to ensure reproducibility and facilitate future research, we plan to release our code and dataset publicly post-publication of the paper. The code and data will be made available on request until publication.

\section{Acknowledgments}
This work was supported by the Cognitive Neuroscience Lab, BITS Pilani, Goa, India. We would like to thank Dr. Rahul Thakur (IoT Lab, Indian Institute of Technology – Roorkee) for providing the computational resources to carry out the experiments.

\bibliographystyle{splncs04}
\bibliography{custom}

\begin{thebibliography}{10}
\providecommand{\url}[1]{\texttt{#1}}
\providecommand{\urlprefix}{URL }
\providecommand{\doi}[1]{https://doi.org/#1}

\bibitem{abnar-etal-2019-blackbox}
Abnar, S., Beinborn, L., Choenni, R., Zuidema, W.: Blackbox meets blackbox:
  Representational similarity {\&} stability analysis of neural language models
  and brains. In: Proceedings of the 2019 ACL Workshop BlackboxNLP: Analyzing
  and Interpreting Neural Networks for NLP. pp. 191--203. Association for
  Computational Linguistics, Florence, Italy (Aug 2019).
  \doi{10.18653/v1/W19-4820}, \url{https://aclanthology.org/W19-4820}

\bibitem{ahmad-etal-2021-unified}
Ahmad, W., Chakraborty, S., Ray, B., Chang, K.W.: Unified pre-training for
  program understanding and generation. In: Proceedings of the 2021 Conference
  of the North American Chapter of the Association for Computational
  Linguistics: Human Language Technologies. pp. 2655--2668. Association for
  Computational Linguistics, Online (Jun 2021).
  \doi{10.18653/v1/2021.naacl-main.211},
  \url{https://aclanthology.org/2021.naacl-main.211}

\bibitem{austin2021program}
Austin, J., Odena, A., Nye, M., Bosma, M., Michalewski, H., Dohan, D., Jiang,
  E., Cai, C., Terry, M., Le, Q., Sutton, C.: Program synthesis with large
  language models (2021). \doi{10.48550/ARXIV.2108.07732},
  \url{https://arxiv.org/abs/2108.07732}

\bibitem{bender-koller-2020-climbing}
Bender, E.M., Koller, A.: Climbing towards {NLU}: {On} meaning, form, and
  understanding in the age of data. In: Proceedings of the 58th Annual Meeting
  of the Association for Computational Linguistics. pp. 5185--5198. Association
  for Computational Linguistics, Online (Jul 2020).
  \doi{10.18653/v1/2020.acl-main.463},
  \url{https://aclanthology.org/2020.acl-main.463}

\bibitem{chen2021codex}
Chen, M., Tworek, J., Jun, H., Yuan, Q., Pinto, H.P.d.O., Kaplan, J., Edwards,
  H., Burda, Y., Joseph, N., Brockman, G., Ray, A., Puri, R., Krueger, G.,
  Petrov, M., Khlaaf, H., Sastry, G., Mishkin, P., Chan, B., Gray, S., Ryder,
  N., Pavlov, M., Power, A., Kaiser, L., Bavarian, M., Winter, C., Tillet, P.,
  Such, F.P., Cummings, D., Plappert, M., Chantzis, F., Barnes, E.,
  Herbert-Voss, A., Guss, W.H., Nichol, A., Paino, A., Tezak, N., Tang, J.,
  Babuschkin, I., Balaji, S., Jain, S., Saunders, W., Hesse, C., Carr, A.N.,
  Leike, J., Achiam, J., Misra, V., Morikawa, E., Radford, A., Knight, M.,
  Brundage, M., Murati, M., Mayer, K., Welinder, P., McGrew, B., Amodei, D.,
  McCandlish, S., Sutskever, I., Zaremba, W.: Evaluating large language models
  trained on code (2021). \doi{10.48550/ARXIV.2107.03374},
  \url{https://arxiv.org/abs/2107.03374}

\bibitem{feng-etal-2020-codebert}
Feng, Z., Guo, D., Tang, D., Duan, N., Feng, X., Gong, M., Shou, L., Qin, B.,
  Liu, T., Jiang, D., Zhou, M.: {C}ode{BERT}: A pre-trained model for
  programming and natural languages. In: Findings of the Association for
  Computational Linguistics: EMNLP 2020. pp. 1536--1547. Association for
  Computational Linguistics, Online (Nov 2020).
  \doi{10.18653/v1/2020.findings-emnlp.139},
  \url{https://aclanthology.org/2020.findings-emnlp.139}

\bibitem{pile}
Gao, L., Biderman, S., Black, S., Golding, L., Hoppe, T., Foster, C., Phang,
  J., He, H., Thite, A., Nabeshima, N., Presser, S., Leahy, C.: The {P}ile: An
  800gb dataset of diverse text for language modeling. arXiv preprint
  arXiv:2101.00027  (2020)

\bibitem{gauthier-levy-2019-linking}
Gauthier, J., Levy, R.: Linking artificial and human neural representations of
  language. In: Proceedings of the 2019 Conference on Empirical Methods in
  Natural Language Processing and the 9th International Joint Conference on
  Natural Language Processing (EMNLP-IJCNLP). pp. 529--539. Association for
  Computational Linguistics, Hong Kong, China (Nov 2019).
  \doi{10.18653/v1/D19-1050}, \url{https://aclanthology.org/D19-1050}

\bibitem{codesearchnet}
Husain, H., Wu, H.H., Gazit, T., Allamanis, M., Brockschmidt, M.: Codesearchnet
  challenge: Evaluating the state of semantic code search (2019).
  \doi{10.48550/ARXIV.1909.09436}, \url{https://arxiv.org/abs/1909.09436}

\bibitem{rsa2008}
Kriegeskorte, N., Mur, M., Bandettini, P.: Representational similarity analysis
  - connecting the branches of systems neuroscience. Frontiers in Systems
  Neuroscience  \textbf{2} (2008). \doi{10.3389/neuro.06.004.2008},
  \url{https://www.frontiersin.org/article/10.3389/neuro.06.004.2008}

\bibitem{rsa2000}
Laakso, A., Cottrell, G.: Content and cluster analysis: Assessing
  representational similarity in neural systems. Philosophical Psychology
  \textbf{13}(1),  47--76 (2000). \doi{10.1080/09515080050002726}

\bibitem{lepori-mccoy-2020-picking}
Lepori, M., McCoy, R.T.: Picking {BERT}{'}s brain: Probing for linguistic
  dependencies in contextualized embeddings using representational similarity
  analysis. In: Proceedings of the 28th International Conference on
  Computational Linguistics. pp. 3637--3651. International Committee on
  Computational Linguistics, Barcelona, Spain (Online) (Dec 2020).
  \doi{10.18653/v1/2020.coling-main.325},
  \url{https://aclanthology.org/2020.coling-main.325}

\bibitem{codexglue}
Lu, S., Guo, D., Ren, S., Huang, J., Svyatkovskiy, A., Blanco, A., Clement, C.,
  Drain, D., Jiang, D., Tang, D., Li, G., Zhou, L., Shou, L., Zhou, L., Tufano,
  M., Gong, M., Zhou, M., Duan, N., Sundaresan, N., Deng, S.K., Fu, S., Liu,
  S.: Codexglue: A machine learning benchmark dataset for code understanding
  and generation (2021). \doi{10.48550/ARXIV.2102.04664},
  \url{https://arxiv.org/abs/2102.04664}

\bibitem{merchant-etal-2020-happens}
Merchant, A., Rahimtoroghi, E., Pavlick, E., Tenney, I.: What happens to {BERT}
  embeddings during fine-tuning? In: Proceedings of the Third BlackboxNLP
  Workshop on Analyzing and Interpreting Neural Networks for NLP. pp. 33--44.
  Association for Computational Linguistics, Online (Nov 2020).
  \doi{10.18653/v1/2020.blackboxnlp-1.4},
  \url{https://aclanthology.org/2020.blackboxnlp-1.4}

\bibitem{puri2021codenet}
Puri, R., Kung, D.S., Janssen, G., Zhang, W., Domeniconi, G., Zolotov, V.,
  Dolby, J., Chen, J., Choudhury, M., Decker, L., Thost, V., Buratti, L.,
  Pujar, S., Ramji, S., Finkler, U., Malaika, S., Reiss, F.: Codenet: A
  large-scale {AI} for code dataset for learning a diversity of coding tasks.
  In: Thirty-fifth Conference on Neural Information Processing Systems Datasets
  and Benchmarks Track (Round 2) (2021),
  \url{https://openreview.net/forum?id=6vZVBkCDrHT}

\bibitem{rogers-etal-2020-primer}
Rogers, A., Kovaleva, O., Rumshisky, A.: A primer in {BERT}ology: What we know
  about how {BERT} works. Transactions of the Association for Computational
  Linguistics  \textbf{8},  842--866 (2020). \doi{10.1162/tacl\_a\_00349},
  \url{https://aclanthology.org/2020.tacl-1.54}

\bibitem{blackbox2022sun}
Sun, T., Shao, Y., Qian, H., Huang, X., Qiu, X.: Black-box tuning for
  language-model-as-a-service. CoRR  \textbf{abs/2201.03514} (2022),
  \url{https://arxiv.org/abs/2201.03514}

\bibitem{wang-etal-2021-codet5}
Wang, Y., Wang, W., Joty, S., Hoi, S.C.: {C}ode{T}5: Identifier-aware unified
  pre-trained encoder-decoder models for code understanding and generation. In:
  Proceedings of the 2021 Conference on Empirical Methods in Natural Language
  Processing. pp. 8696--8708. Association for Computational Linguistics, Online
  and Punta Cana, Dominican Republic (Nov 2021).
  \doi{10.18653/v1/2021.emnlp-main.685},
  \url{https://aclanthology.org/2021.emnlp-main.685}

\bibitem{yu-etal-2018-spider}
Yu, T., Zhang, R., Yang, K., Yasunaga, M., Wang, D., Li, Z., Ma, J., Li, I.,
  Yao, Q., Roman, S., Zhang, Z., Radev, D.: {S}pider: A large-scale
  human-labeled dataset for complex and cross-domain semantic parsing and
  text-to-{SQL} task. In: Proceedings of the 2018 Conference on Empirical
  Methods in Natural Language Processing. pp. 3911--3921. Association for
  Computational Linguistics, Brussels, Belgium (Oct-Nov 2018).
  \doi{10.18653/v1/D18-1425}, \url{https://aclanthology.org/D18-1425}

\end{thebibliography}
\end{document}